\documentclass[11pt]{article}
\usepackage{geometry}
\usepackage{coling2020}
\usepackage{times}
\usepackage{url, hyperref}
\usepackage{latexsym}
\usepackage{microtype}
\usepackage{textcomp}
\usepackage{marvosym}
\usepackage{pgfplots}
\usepackage{booktabs} % pretty tables
\usepackage{array}
\usepackage{standalone} % import figures
\usepackage{tikz}
\usepackage{siunitx} % align floating point numbers in tables
\usetikzlibrary{positioning,fit,calc,arrows.meta,backgrounds}
\usepackage{svg}
\usepackage{amsmath}
\usepackage{caption}
\usepackage[title]{appendix}
\usepackage{etoolbox}
\apptocmd{\appendices}{\apptocmd{\thesection}{.}{}{}}{}{}

\pgfplotsset{compat=newest}
\usepgfplotslibrary{colorbrewer}

\colingfinalcopy % Uncomment this line for all SemEval submissions

\title{CyberWallE at SemEval-2020 Task 11: An Analysis of Feature Engineering for Ensemble Models for Propaganda Detection}
\author{Verena Blaschke\qquad Maxim Korniyenko\qquad Sam Tureski\\
Seminar f\"{u}r Sprachwissenschaft\\
Eberhard Karls Universit\"{a}t T\"{u}bingen\\
{\tt first.last@student.uni-tuebingen.de}}

\date{}

\begin{document}
\maketitle
\blfootnote{This work is licensed under a Creative Commons Attribution 4.0 International Licence. \\
Licence details: \url{http://creativecommons.org/licenses/by/4.0/}.}
\begin{abstract}
% 1) General info about the paper.
This paper describes our participation in the SemEval-2020 task Detection of Propaganda Techniques in News Articles.
% 2) Mention that we participated in both tasks and give both full names of those tasks and their abbreviations if we use them again in the abstract. It's ok if this part is repeated again in the main part of the article.
We participate in both subtasks: Span Identification (SI) and Technique Classification (TC).
% 3) Give a very brief description of both of our architectures. Issues 2 and 3 could be put together into one sentence.
We use a bi-LSTM architecture in the SI subtask and train a complex ensemble model for the TC subtask.
Our architectures are built using embeddings from BERT in  combination with additional lexical features and extensive label post-processing.
% 4) Mention the results of both of our models. (Two digits after the dot given our current notation for numbers).
Our systems achieve a rank of 8 out of 35 teams in the SI subtask (F1-score: 43.86\%) and 8 out of 31 teams in the TC subtask (F1-score: 57.37\%).
% ?) Provide additional info about the performance of the structure\technique\technology. 
%    Not sure that we should include here something.
%    If we write here something we could mention that our experiments show that BERT embeddings outperform other embeddings.
\end{abstract}

\section{Introduction}

% What is the task about and why is it important? Be sure to mention the language(s) covered and cite the task overview paper. ~1 paragraph

% What is the main strategy your system uses? ~1 paragraph

% What did you discover by participating in this task? Key quantitative and qualitative results, such as how you ranked relative to other teams and what your system struggles with. ~1 paragraph

% Have you released your code? Give a URL

%Alternative quotes:
%The Canadian academic Douglas N. Walton argues that ``the aim of propaganda is to get the respondent to act, to adopt a certain course of action, or to go along with and assist in a particular policy" \cite{walton2007media}.
Propaganda is defined as ``the deliberate and systematic attempt to shape perceptions, manipulate cognitions, and direct behavior to achieve a response that furthers the desired intent of the propagandist" \cite[p.~6]{jowett2019propaganda}.
% The political scientist Harold Laswell defined propaganda as ``the expression of opinions or actions carried out deliberately by individuals or groups with a view to influencing the opinions or actions of other individuals or groups for predetermined ends and through psychological manipulations" \cite{ellul1973propaganda}.
With the advent of rapid dissemination of news articles through online social media, automatic detection of biased or fake reporting has become more crucial than ever before.

This paper describes our participation in both of the subtasks offered by \newcite{DaSanMartinoSemeval20task11} in the SemEval 2020 shared task for the Detection of Propaganda Techniques in News Articles. The Span Identification (SI) subtask is a binary classification problem to discover propaganda at the token level, and the Technique Classification (TC) subtask involves a 14-way classification of propagandistic text fragments. 

To address the SI task, we combine token-level BERT embeddings and linguistic features with bidirectional LSTMs and data post-processing methods. 
To address the TC task, we use BERT sequence embeddings and linguistic features to train a feed-forward neural network before post-processing is applied. 

While top-scoring teams in a similar shared task \cite{EMNLP19DaSanMartino}  focus primarily on leveraging the performance of the pre-trained, context-dependent language model BERT, we find that further encoding of linguistic features offers a meaningful boost over both BERT and GloVe word embeddings alone.
Majority voting and span merging in the SI task, as well as pre- and post-processing techniques to account for frequent occurrences of the \textsc{repetition} technique in the TC task result in further performance increases.
Along with our best-performing model, we provide an extensive exploration into various embedding, feature and classifier combinations. 
% Significant drops in performance on final test set data for nearly all teams in the 2020 shared task suggest that robustness against inconsistencies in news data quality, class distribution, and propaganda annotation should remain an important topic of further research.

We release our code at \href{https://github.com/cicl-iscl/CyberWallE-propaganda-detection}{\texttt{github.com/cicl-iscl/CyberWallE-propaganda-detection}}.

\section{Background}

Before the introduction of the 2019 shared task on propaganda detection, approaches to propaganda recognition in news generally focused on classification at the article level.
\newcite{rashkin2017truth} compare the language of ``trusted news articles" from a well-known corpus with stories from ``unreliable news sites". The publicly-available, feature-based tool Proppy, released by \newcite{barron2019proppy}, ranks articles by their likelihood of containing propagandist content. These previous systems have been prone to misclassification and lack of explainability, however, due to the assumption that articles from news sources deemed propagandist will always contain propaganda.

The release of the shared task on Fine-Grained
Propaganda Detection by \newcite{da2019findings} sparked the creation of numerous new detection systems, resulting in 14 system papers published. In contrast to this year's tasks, the 2019 participants encountered sentences, rather than tokens, in the binary classification round, and were asked to differentiate 18, rather than 14 classes, in the multiclass classification round. 

The highest-scoring teams in that task integrate the language representation model BERT (Bidirectional Encoder Representations from Transformers) into their systems.
BERT, developed by \newcite{devlin2019bert}, was shown to reach state-of-the-art performance on eleven natural language processing tasks at the time of its release. 
The \texttt{base} version of the model for English contains 110 million parameters over 12 layers, produces embeddings of size 768 and encodes semantic information on a sub-token level. 

% A number of the previous year's teams demonstrate success by including linguistic features in addition to BERT or other word embeddings. 
A number of the previous year's teams use linguistic features in addition to BERT or other word embeddings. 
% \newcite{gupta-etal-2019-neural} employ six categories of features, ranging from LDA-based topical representations to layout-derived features like sentence length.
\newcite{gupta-etal-2019-neural} employ six categories of features, ranging from topic representations to layout-derived features like sentence length. 
% \newcite{ferreira-cruz-etal-2019-sentence} apply simple linguistic features (e.g. punctuation frequency, sentence length) to ELMo embeddings, with the notable additions of type-token ratios and TF-IDF scores for uni- and bigrams. 
\newcite{ferreira-cruz-etal-2019-sentence} work with simple linguistic features (e.g. punctuation frequency, sentence length) as well as type-token ratios and TF-IDF scores for uni- and bigrams. 
\newcite{al-omari-etal-2019-justdeep}, by contrast, employ the output of a twitter-based sentiment analysis tool. % in addition to their base word representations.
%All three of these teams also use ensemble methods. (ADD THIS or something about their classifiers?)
Finally, \newcite{alhindi-etal-2019-fine} and \newcite{li-etal-2019-detection} both use features returned by the LIWC (Linguistic Inquiry and Word Count) text analysis software \cite{pennebaker2001linguistic}. 

\section{Dataset}\label{dataset}
The corpus used is an extension of the corpus described in \newcite{EMNLP19DaSanMartino}. 
The articles, pulled from around 50 news outlets, have been annotated into specific techniques at the fragment level; several classes with few instances are then condensed under single labels, such as in \textsc{whataboutism, straw men, red herring}. 
An example of a fragment containing an instance of the class \textsc{loaded language}:

\vspace{-0.7em}
{
\centering
$\text{\small not} \overbrace{\text{\small looking as though Trump killed his grandma}}^\text{loaded language}\text{\small.}$
\\
}
Descriptions of each class can be found in \newcite{DaSanMartinoSemeval20task11}. 
%don't get into sentence length details most frequent classes are the shortest on average/ tend to have shorter spans, roughly on average 3-4 tokens, even longest class appeal to authority has 24 tokens on average

% \subsection{Class distribution}
Class imbalance exists strongly in the dataset. 
A minority of sentences are assessed to contain propaganda, and fragments that do contain propaganda are classed as only 3 of the 14 labels around 60\% percent of the time. 
It is important to note that some of the techniques can be identified on the fragment level  (e.g. \textsc{name calling, flag-waving}), 
while others often require a broader context of the discourse at hand to be discovered (e.g. \textsc{repetition, red herring}).

% \subsection{Inconsistencies in labelling}

% Upon inspection of the gold-standard training set labels, we find that labelling inconsistencies or omissions may exist in the dataset, especially for the \textsc{repetition} class. For example, the term ”Countering Violent Extremism” occurs at least 4 times in one article (785801366), yet is only labelled each time as a \textsc{slogan}. In another article (771546417), the synonymous fragments “Make America Great Again” and “MAGA” occur 4 times; 3 times labeled as a \textsc{slogan} and the final time as \textsc{repetition}. 
% Whether this is a result of the inter-annotator disagreement noted by Da San Martino et al. (2019b) in the corpus description paper remains to be known. 

% Furthermore, labelling for \textsc{slogans} and \textsc{flag\_waving} may also be difficult to discern. Article \#729668796 features the terms “Hungary First, ”For us,
% Hungary is first,” and “America First” as both \textsc{slogan} and \textsc{flag\_waving} each time.

\section{The Span Identification (SI) system}

% Replicability: present all details that will allow someone else to replicate your system

% Key algorithms and modeling decisions in your system; resources used beyond the provided training data; challenging aspects of the task and how your system addresses them. This may require multiple pages and several subsections, and should allow the reader to mostly reimplement your system’s algorithms.

% Use equations and pseudocode if they help convey your original design decisions, as well as explaining them in English. If you are using a widely popular model/algorithm like logistic regression, an LSTM, or stochastic gradient descent, a citation will suffice—you do not need to spell out all the mathematical details.

% Give an example if possible to describe concretely the stages of your algorithm.

% If you have multiple systems/configurations, delineate them clearly.

% This is likely to be the longest section of your paper.

\begin{figure}
\begin{minipage}[b]{.38\textwidth}
\captionsetup{}
  \vspace*{\fill}
  \centering
    \definecolor{inputcol}{RGB}{240, 248, 255}
\definecolor{featurecol}{RGB}{74, 186, 186}
\definecolor{modelcol}{RGB}{252, 240, 146}
\definecolor{labelcol}{RGB}{250, 127, 77}
\tikzset{
    base/.style = {font=\small},
    box/.style = {base, rectangle, rounded corners, draw=black,
                   minimum width=0.1cm, minimum height=0.5cm, 
                   text centered, node distance=1.2cm},
    input/.style = {box, fill=inputcol},
    lstm/.style = {box, minimum width=1.5cm, fill=modelcol!80, node distance=1.7cm},
    lstm2/.style = {lstm, fill=modelcol!70, draw=black!70},
    lstm3/.style = {lstm, fill=modelcol!60, draw=black!60},
    lstm4/.style = {lstm, fill=modelcol!50, draw=black!50},
    lstm5/.style = {lstm, fill=modelcol!40, draw=black!40},
    feature/.style = {box, fill=featurecol!50, 
                      minimum height=9mm, node distance=1.5cm},
    group/.style = {box, fill=featurecol!30},
    labels/.style = {box, minimum width=1.5cm, fill=labelcol!90},
    labels2/.style = {labels, fill=labelcol!75, draw=black!70},
    labels3/.style = {labels, fill=labelcol!65, draw=black!60},
    labels4/.style = {labels, fill=labelcol!55, draw=black!50},
    labels5/.style = {labels, fill=labelcol!45, draw=black!40},
    bg2/.style = {black!70},
    bg3/.style = {black!60},
    bg4/.style = {black!50},
    bg5/.style = {black!40},
    >=LaTeX
}
\begin{tikzpicture}[
    node distance=1.5cm,
    align=center % necessary for intra-node linebreaks
    ]
%   \selectcolormodel{gray} % for checking readability in grayscale
  \node (senti)      [feature]   {sentiment\\features};
  \node (bert)     [feature, left=1mm of senti]          {BERT\\embedding};
  \node (arglex)      [feature, right=1mm of senti]   {rhetorical\\feature};
  \node (pos)      [feature, right=1mm of arglex]   {POS\\tag};
  \begin{scope}[on background layer]
      \node (features) [group, fit={(bert) (senti) (arglex) (pos)}] {};
  \end{scope}
  \node (input)             [input, below of=features]              {sentence-wise input tokens};
  \node (lstm) [lstm, above of=features] {bi-LSTM};
   \begin{scope}[on background layer]
      \node (lstm5) [lstm5, below right=-1mm and -6mm of lstm] {};
      \node (lstm4) [lstm4, above left=-4mm and -13mm of lstm5] {};
      \node (lstm3) [lstm3, above left=-4mm and -13mm of lstm4] {};
      \node (lstm2) [lstm2, above left=-4mm and -13mm of lstm3] {};
  \end{scope}
  
  \node (labels5) [labels5, above of=lstm5] {};
  \node (labels4) [labels4, above of=lstm4] {};
  \node (labels3) [labels3, above of=lstm3] {};
  \node (labels2) [labels2, above of=lstm2] {};
  \node (labels) [labels, above of=lstm] {IO labels};
  
  \node (majlabels) [labels, above of=labels] {consolidated IO labels};
  \node (rawspans) [labels, above of=majlabels] {raw spans};
  \node (finalspans) [labels, above of=rawspans] {final spans};
  
  \draw[->]             (input) -- (bert.south east);
  \draw[->]             (input) -- (senti.south);
  \draw[->]             (input) -- (arglex);
  \draw[->]             (input) -- (pos.south west);
  
  \draw[->, bg2]             (features.north) -- (lstm2);
  \draw[->, bg3]             (features.north) -- (lstm3);
  \draw[->, bg4]             (features.north) -- (lstm4);
  \draw[->, bg5]             (features.north) -- (lstm5);
  
  \draw[->]             (lstm) -- (labels);
  \draw[->, bg2]             (lstm2) -- (labels2);
  \draw[->, bg3]             (lstm3) -- (labels3);
  \draw[->, bg4]             (lstm4) -- (labels4);
  \draw[->, bg5]             (lstm5) -- (labels5);
  
  \draw[->, bg5]             (labels5) -- (majlabels.south);
  \draw[->, bg4]             (labels4) -- (majlabels.south);
  \draw[->, bg3]             (labels3) -- (majlabels.south);
  \draw[->, bg2]             (labels2) -- (majlabels.south);
  \draw[->]             (labels) -- node[base, left] {majority voting} (majlabels.south);
  
  \draw[->]             (majlabels) -- node[base, left] {conversion to spans} (rawspans);
  \draw[->]             (rawspans) -- node[base, left] {merging close spans} (finalspans);
  
  % Drawing these again, so the layering works:
  \node[lstm2, above left=-4mm and -13mm of lstm3] {};
  \node[lstm, above of=features] {bi-LSTM};
  \draw[->]             (features) -- (lstm);
  \node (labels2) [labels2, above of=lstm2] {};
  \node (labels) [labels, above of=lstm] {IO labels};
  \draw[->]             (lstm) -- (labels);
  \end{tikzpicture}
  \caption{\small{The span identification system.}}
  \label{fig:task1-architecture}
  \end{minipage}
\begin{minipage}[b]{.5\textwidth}
\captionsetup{}
  \vspace*{\fill}
  \centering
\definecolor{inputcol}{RGB}{240, 248, 255}
\definecolor{featurecol}{RGB}{74, 186, 186}
\definecolor{modelcol}{RGB}{252, 240, 146}
\definecolor{labelcol}{RGB}{250, 127, 77}
\tikzset{
    base/.style = {font=\small}, % \small\sffamily
    box/.style = {base, rectangle, rounded corners, draw=black,
                   minimum width=0.1cm, minimum height=0.5cm, 
                   text centered},
    input/.style = {box, fill=inputcol},
    lstm/.style = {box, minimum width=1.5cm,fill=modelcol!80},
    feature/.style = {box, fill=featurecol!50,
                   minimum height=1cm,},
    group/.style = {box, fill=featurecol!30},
    labels/.style = {box, minimum width=1.5cm, fill=labelcol!90},
    bg2/.style = {black!70},
    bg3/.style = {black!60},
    bg4/.style = {black!50},
    bg5/.style = {black!40},
    >=LaTeX
}
\begin{tikzpicture}[
    node distance=1.5cm,
    align=center % necessary for intra-node linebreaks
    ]
%   \selectcolormodel{gray} % for checking readability in grayscale
  \node (input)             [input]              {text fragments};
  \node (input-rep) [input, above right=3mm and 3mm of input] {text fragments\\with repetition\\pre-processing};
  \node (input-wo-rep) [input, right=3mm of input-rep] {text fragments\\w/o repetition\\instances};
  \node (input-dup) [input, above left=2.5cm and 1.2cm of input] {text fragments\\with repetition\\+ duplicate\\annotation};
  
  \node (bert)     [feature, above of=input-rep]          {BERT\\embeddings};
  \node (lin) [lstm, above=4mm of bert] {linear\\classifier};
  
  \node (presoft) [feature, above=4mm of lin] {pre-soft-\\max layer};
  
  \node (question)      [feature, left=1mm of presoft]   {question\\feature};
  \node (ner)      [feature, left=1mm of question]   {NE\\features};
  \node (arglex)      [feature, left=1mm of ner]   {rhetorical\\features};
  \begin{scope}[on background layer]
      \node (features) [group, fit={(presoft) (arglex) (question) (ner)}] {};
  \end{scope}
  
  \node (bert2)     [feature, above of=input-wo-rep]          {BERT\\embeddings};
  
  \node (lin2) [lstm, above=4mm of bert2] {linear\\classifier};
  \node (ffnn) [lstm, above of=features] {multilayer\\perceptron};
  
  \node (ffnn-labels) [labels, above=4mm of ffnn] {labels};
  \node (labels-wo-rep) [labels, above=2.2cm of lin2] {labels w/o\\repetition};
  
  \node (labels) [labels, above of=ffnn-labels] {labels};

  \draw[->]             (input) -- (input-rep);
  \draw[->]             (input) -- (input-wo-rep.south west);
  \draw[->]             (input) -- (input-dup);

  \draw[->]             (input-rep) -- (bert.south);
  \draw[->]             (input-wo-rep) -- (bert2.south);
  
  \draw[->]             (input) -- (arglex.south east);
  \draw[->]             (input) -- (ner);
  \draw[->]             (input) -- (question.south);

  \draw[->]             (bert) -- (lin);
  \draw[->]             (bert2) -- (lin2);
  
  \draw[->]             (lin) -- (presoft);
  
  \draw[->]             (features) -- (ffnn);
  
  \draw[->]             (ffnn) -- (ffnn-labels);
  \draw[->]             (lin2) -- (labels-wo-rep);
  \draw[->]             (labels-wo-rep) -- (labels);
  \draw[->]             (ffnn-labels.north) -- node[base, fill=white] {label post-processing} (labels);

  \begin{scope}[on background layer]
    \draw[->]             ([xshift=-1cm]input-dup.north) -- (labels);
    % Drawing this again, so the layering works:
    \node (features) [group, fit={(presoft) (arglex) (question) (ner)}] {};
  \end{scope}
  \end{tikzpicture}
  \caption{\small{The technique classification system.}}
  \label{fig:task2-architecture}
\end{minipage}
%\caption{Our system architectures.}
%\label{fig:system-architectures}
\end{figure}

% \begin{figure}
% \centering
% \begin{subfigure}[b]{.38\textwidth}
%     \centering
%     \includestandalone[width=\textwidth]{figures/task1-system}
%     \caption{The span identification system.}
%     \label{fig:task1-architecture}
% \end{subfigure}
% \begin{subfigure}[b]{.61\textwidth}
%     \centering
%     \includestandalone[width=\textwidth]{figures/task2-system}
%     \caption{The technique classification system.}
%     \label{fig:task2-architecture}
% \end{subfigure}
% \caption{Our system architectures.}
% \label{fig:system-architectures}
% \end{figure}
\subsection{System overview}

Our architecture for the SI subtask is built around a bidirectional LSTM \cite{graves2005framewise}.
We split the news articles into sentences (or sentence fragments) and tokenize these sentences. We convert the target spans into corresponding token-wise binary labels (\textit{I} - inside a span, \textit{O} - outside a span).

Each token is represented as a vector consisting of a
BERT embedding \cite{devlin2019bert} concatenated with two sentiment values (indicating how positive/negative the token is), a binary feature indicating whether the token is part of a rhetorically salient phrase, and a one-hot encoded POS tag representation.
This input is fed into a bidirectional LSTM that predicts a label for each token. 
We carry this out five times in total to abstract over the effect caused by random initializations.
The five sets of predicted labels are consolidated via majority voting and then converted into spans.
In a final step, we remove short gaps between spans by merging the two surrounding spans into one larger span.

Figure \hyperref[fig:task2-architecture]{1} illustrates this architecture.

\subsection{Experimental setup}

We use the Natural Language Toolkit %\footnote{\url{https://www.nltk.org/}}
3.2.4 \cite{bird2009natural}
for sentence splitting and additional heuristics to divide long sentences into shorter fragments by splitting them along quotation marks, semicolons and commas when possible. The maximum sentence length is 35 tokens (long enough to fully include most input fragments, but short enough to reduce vanishing gradient issues).
We use spaCy's pretrained model for English (version 2.0.0) as tokenizer.%\footnote{\url{https://spacy.io/}}

We use HuggingFace's pretrained BERT embeddings \cite{Wolf2019HuggingFacesTS},
%\footnote{\url{https://github.com/huggingface/transformers}}
specifically the uncased base version. 
We initially also experiment with
the cased version, as well as small (6B tokens, 100 dimensions) and large (42B tokens, 300 dimensions) GloVe embeddings \cite{pennington2014glove}.%\footnote{\url{https://nlp.stanford.edu/projects/glove/}}

The sentiment features are based on SentiWordNet 3.0.0 \cite{baccianella2010sentiwordnet}.\footnote{\url{https://github.com/aesuli/SentiWordNet}}
Each token is associated with a positive and a negative sentiment score, each bounded between $0$ and $1$.
Tokens that are not in the dataset (even in their lemmatized form as produced using spaCy's lemmatizer)  are assigned the value $[0, 0]$. 
Polysemous tokens that are assigned several different scores by SentiWordNet are assigned the average of these values.
% (preliminary experiments also with SentiWords \newcite{gatti2015sentiwords} only one score)

The feature for rhetorically salient tokens is produced using the Arguing Lexicon\footnote{\url{http://mpqa.cs.pitt.edu/lexicons/arg_lexicon/}} \cite{somasundaran2007detecting},
which contains phrase patterns for 17 different rhetorical strategies such as \textit{appeal to authority} or \textit{generalization}.
Each token in a phrase contained in this lexicon is assigned the value~$1$, all others~$0$.

The last feature is each token's one-hot encoded 15-dimensional POS tag, as determined by spaCy.

We use Keras %footnote{\url{https://keras.io}} 
2.2.5 with a Tensorflow 1.14 backend to build the LSTM. 
We work with a batch size of 128, class weights (1.0 for \textit{O}, 6.5 for \textit{I}), 512 hidden units and a dropout rate of 25\%. 
We use the Adam optimizer (learning rate: 0.001) to minimize the cross entropy across ten epochs.

After predicting the labels and transforming them into spans, we merge all spans that are separated by less than 25 characters.
We also experiment with using minimum gap lengths of up to fifty characters.

During the development stage, we use
all of the training data as training input to tune parameters based on development data performance.
For the test set submissions, we extract the development data labels from the input for the TC task and concatenate the training and development data as model input.

The predictions are evaluated using character-level precision, recall and F1-score for the propagandist fragments.
More details can be found in the task description paper \cite{DaSanMartinoSemeval20task11}.
% While the TC task  %Task 2
% is an example of multinomial classification, one of the classes stands out in particular. %"is qualitatively different from the others" Perhaps mention that it was important because it was large/contributed a great deal to the micro-averaged F1-score? 
% Most of the instances of the "Repetition" class rely on data that are not encoded in the span itself.
% Our key innovation in resolving this issue is to break the process of generating predictions into several sub-models.

\section{The Technique Classification (TC) system}

% How data splits (train/dev/test) are used.

% Key details about preprocessing, hyperparameter tuning, etc. that a reader would need to know to replicate your experiments. If space is limited, some of the details can go in an Appendix.

% External tools/libraries used, preferably with version number and URL in a footnote.

% Summarize the evaluation measures used in the task.

% You do not need to devote much—if any—space to discussing the organization of your code or file formats.

\subsection{System overview}
The foundation of our strategy for the TC subtask revolves around generating accurate predictions for the large and qualitatively unique \textsc{repetition} class.  
Our key innovation in resolving this issue is to %craft/implement an ensemble system containing several sub-models
break prediction generation into several sub-models.

For the \textit{base model}, we use lexical information encoded in each span to generate a first set of predictions (see the central branch of Figure \hyperref[fig:task2-architecture]{2}).
We modify the input fragments slightly to represent repetition information.
If a fragment has the \textsc{repetition} label (in the case of training data) or is repeated elsewhere in the same news article (development/test data), we append a copy of this sequence to itself.
Then, we train a linear classifier on all fragments, which are represented using BERT embeddings.
Afterwards, we take the pre-softmax layer from this model and concatenate it with additional features
indicating whether a rhetorical technique is contained in the fragment, whether it mentions geopolitical entities or groups of people (two binary features), and whether it contains a question mark.
These newly created vectors are used as input to a multilayer perceptron, which generates this model's final predictions.

For our \textit{repetition model}, we isolate two simple features which are then assigned to every instance of the data---how often the segment is repeated in the article and whether it is the first such repetition (the left branch of Figure \hyperref[fig:task2-architecture]{2}).
This post-processing step classifies instances as \textsc{repetition} if the given normalized span has at least one repetition in the article and it does not represent the first occurrence of it in the text.
These predictions override the base model's predictions.

If the base model predicts a \textsc{repetition} that is not confirmed by this post-processing step, this instance is re-classified by a third model, which we call the \textit{alternative model}
(the right branch of Figure \hyperref[fig:task2-architecture]{2}).
It is a linear classifier which uses BERT embeddings of the data as input and is trained using all training data except the instances of the \textsc{repetition} class.
As a result, the model always predicts labels of only the remaining thirteen classes.

The last step of this system is handling duplicates.
Some of the fragments have several labels and appear as multiple identical instances in the data.
We use the prediction from the alternative model to label the duplicate instance.
If these predictions are identical or if there are three or more identical spans, we use the first model's runner-up predictions instead.

\subsection{Experimental setup}

The \textit{base model} and \textit{alternative model} use HuggingFace's pretrained \texttt{bert-base-uncased}
vectors with a sequence classification head on top, \texttt{BertForSequenceClassification}. The maximum sequence length is 200 tokens, as determined by HuggingFace's pretrained BERT tokenizer.
Our optimizer is \texttt{BertAdam} with a warmup rate of 0.1. 
The base model is trained for two epochs with a learning rate of 1e-5 and a batch size of 12.
The alternative model is trained for four epochs with a learning rate of 2e-5 and a batch size set to 32.
This settings are based on the parameters recommended in \newcite{devlin2019bert}. 

The rhetorical technique feature is generated using the Arguing Lexicon \cite{somasundaran2007detecting}, as in the previous task.
We create the two named entity features using spaCy's pretrained named entity tagger (version 2.0.0) based on its predictions for `NORP' (nationalities, religious or political groups) and `GPE' (countries, cities, states).
% All of these features are binary, indicating the presence or absence of a

In order to build the base model's feed-forward neural network, we use Keras 2.2.5 for Tensorflow 1.14.
The best results on the development set are achieved using a hidden layer with 128 units and a dropout rate of 25\%.
This multilayer perceptron is trained for 15 epochs using a batch size of 128 instances.
To minimize the cross entropy, we use the Adam optimizer with a learning rate set to 0.001.

This final base model is chosen after a series of experiments exploring different input representations and machine learning models.

As alternative BERT embeddings, we also test the cased version of HuggingFace's pretrained base model.
Additionally, we experiment with alternatives to the pre-softmax layer of \texttt{BertForSequenceClassification}.
To embed a sequence with BERT, it needs to be pre- and suffixed with two special embeddings, \texttt{[CLS]} and \texttt{[SEP]}.
BERT's embedding for \texttt{[CLS]} in its final layer represents the entire sequence \cite{devlin2019bert}.
We extract this embedding and concatenate it with the final-layer representations of the first ten actual tokens and feed this input to three different machine learning architectures.
We use the Keras 2.2.5 implementation of a convolutional neural network (CNN) (3 layers of size 128, max pooling, dense layer)  as well as the KimCNN (a CNN architecture developed for NLP tasks by \newcite{kim-2014-convolutional}) and a multilayer-perceptron (same configuration as for our actual base model).

In a further set of experiments, we compare different classifiers on top of the pre-softmax layer from \texttt{BertForSequenceClassification}.
Firstly, we inspect the actual output of \texttt{BertForSequenceClassification}.
We also test out the following classifiers: a decision tree with extreme gradient boosting \cite{chen2016xgboost}, a single-layer perceptron (with the same set-up as the neural net in the base model, only without the hidden layer) and a support vector machine (SVM) with the default configuration of Scikit-learn version 0.22.2 \cite{scikit-learn}.

We also explore different input features (and combinations thereof) in an architecture otherwise identical to our actual base model.
We use four additional binary named entity features (also created with spaCy) that indicate whether a text fragment contains organizations, names of people, cardinal numbers, and dates.
Moreover, we test binary features indicating whether a fragment contains tokens related to the US (\textit{America}) or to the \textsc{reductio ad hitlerum} class (\textit{reductio}).
Furthermore, we use the Natural Language Understanding Tool by IBM Cloud\footnote{\url{https://cloud.ibm.com/catalog/services/natural-language-understanding}} to get each fragment's anger, disgust, fear, joy and sadness ratings.
Each value is between $0.0$ and $1.0$ and stored as an individual feature.
We also examine two numerical features, one encoding the sequence length (in tokens) and the other encoding how often a text fragment is repeated in the given news article.

This subtask is scored based on micro-averaged F1-score.

\section{Results}

% Main quantitative findings: How well did your system perform at the task according to official metrics? How does it rank in the competition?

% Quantitative analysis: Ablations or other comparisons of different design decisions to better understand what works best. Indicate which data split is used for the analyses (e.g. in table captions). If you modify your system subsequent to the official submission, clearly indicate which results are from the modified system.

% Error analysis: Look at some of your system predictions to get a feel for the kinds of mistakes it makes. If appropriate to the task, consider including a confusion matrix or other analysis of error subtypes—you may need to manually tag a small sample for this.

% Analysis: focus more on results and analysis and less on discussing rankings; report results on several runs of the system (even beyond the official submissions); present ablation experiments showing usefulness of different features and techniques; show comparisons with baselines.

\begin{table}[t]
    \centering
    \renewcommand\arraystretch{1.1}
\newcolumntype{R}{>{\raggedleft\arraybackslash}p{1.5cm}}
\begin{tabular}{@{}lRRR@{}}
    \toprule
    \textbf{Configuration} & \textbf{F1-score} & \textbf{Precision} & \textbf{Recall} \\
    \midrule
    GloVe-6B-100D (uncased) & 0.3148 & 0.2015  & 0.7307 \\
    GloVe-6B-100D  + SentiWordNet (\textsc{\textsc{swn}}) & 0.3171 & 0.2052 & 0.7059 \\
    GloVe-6B-100D  + Arguing Lexicon (\textsc{al}) & 0.3105 &	0.1989 & \textbf{0.7359}\\
    GloVe-6B-100D  + \textsc{swn} + \textsc{al} & 0.3223 & 0.2097 & 0.7026\\
    GloVe-6B-100D  + \textsc{pos} & 0.3294 & 0.2169 & 0.6897\\
    GloVe-6B-100D  + \textsc{swn} + \textsc{al} + \textsc{pos} & 0.3247 &	0.2132 & 0.6875\\
    % \rule{0pt}{3ex}%
    \hline
    GloVe-42B-300D (uncased) & 0.3453 & 0.2472  & 0.5792 \\
    % \rule{0pt}{3ex}%
    \hline
    \textsc{bert}-base-cased &  0.3759  & 0.2792  &  0.5783 \\
    % \rule{0pt}{3ex}%
    \hline
    \textsc{bert}-base-uncased        & 0.3869 & 0.2884  & 0.5965 \\
    \textsc{bert}-base-uncased + \textsc{swn} & 0.3893 &	0.2966 &	0.5784 \\
    \textsc{bert}-base-uncased + \textsc{al} & 0.3921 & 0.3066	& 0.5507\\
    \textsc{bert}-base-uncased + \textsc{swn} + \textsc{al} & 0.3912 & 0.2995 & 0.5687\\
    \textsc{bert}-base-uncased + \textsc{pos} & 0.3912 &	0.2939 & 0.5915\\
    \textsc{bert}-base-uncased + \textsc{swn} + \textsc{al} + \textsc{pos} ~~(``full'') & 0.3949 & 0.3025 & 0.5699 \\
    % \rule{0pt}{3ex}%
    \hline
    full + majority voting (across 5 initializations) & 0.4065 & 0.3110 & 0.5901 \\
    \textit{full + majority voting + span merging} &  \textbf{0.4115} & \textbf{0.3136} & 0.6026\\
    \bottomrule
\end{tabular}

    \caption{Different embeddings and feature combinations for the development set of the SI task. The results are mean values across five runs. The configuration for our final model is in italics.}
    \label{tab:task1-ablation}
\end{table}

\subsection{Span identification}

Our best model achieves an F1-score of 42.39\% on the development set and 43.86\% on the test set (rank 8 of 35).
In this subsection, we describe the findings of our feature ablation experiments.
The detailed results can be found in Table~\ref{tab:task1-ablation}.

Using the simplest token embedding (100-dimensional GloVe) yields a low precision but very high recall score (20.15\% compared to 73.07\%).
Using larger (300-dimensional GloVe) or more sophisticated embeddings (BERT) lowers the system's recall score while increasing precision enough to also increase the F1-score.
% does uncased count as larger? (-> bert)
Adding linguistic features generally has similar effects.
It should be noted, however, that while adding features mostly raises the overall score,
the strength of this effect and the best composition of feature combinations vary greatly across runs.

% Adding linguistic features to the GloVe embeddings has a slightly larger effect than adding them to the BERT embeddings.
% -- actually really not that much (0.01 vs. 0.008), so skipping this

We observe a significant increase in performance by adding post-processing steps that make the model both more robust to initialization differences and improve the predictions.
Majority voting across 5 model initializations raises the F1-score by more than one percentage point.
During almost all runs, we observe that the predictions after majority voting have higher precision and recall scores than any of the individual predictions that went into this voting process.
Merging nearby spans also yields better results for both metrics.
We achieve good results for a range of minimum gap lengths (between 10 and 40 characters); optimum values within this range do not generalize between model initializations.

% Are there spans that don't follow token boundaries???

\begin{table}[t]
    \centering
    \renewcommand\arraystretch{1.1}
\newcolumntype{R}{>{\raggedleft\arraybackslash}p{1.5cm}}
\begin{tabular}{@{} l R @{}}
\toprule
\textbf{Configuration}     & \textbf{F1-score}     \\
\midrule
\textsc{bert}-base-cased + pre-softmax + multilayer perceptron (MLP) & 0.5485\\
\textsc{bert}-base-uncased + pre-softmax + MLP & 0.5635
\\ % NO REPETITION PREPROCESSING
% \rule{0pt}{3ex}%
\hline
\textsc{bert}-base-uncased embeddings of \texttt{[CLS]} \& the first 10 tokens + MLP & 0.5870\\
% \rule{0pt}{3ex}%
% \cmidrule[0.02em]{1-2}
\hline
rep + \textsc{bert}-base-uncased + pre-softmax + MLP & 0.6341 \\
% \rule{0pt}{3ex}%
\hline
rep + \textsc{bert}-base-uncased + pre-softmax + two named entity classes (\textsc{ne-2}) + MLP  & 0.6322\\ 
rep + \textsc{bert}-base-uncased + pre-softmax + Arguing Lexicon (\textsc{al}) + MLP & 0.6350 \\
rep + \textsc{bert}-base-uncased + pre-softmax + question mark feature (\textsc{q}) + MLP  & 0.6359\\
rep + \textsc{bert}-base-uncased + pre-softmax + \textsc{ne-2} + \textsc{al} + \textsc{q} + MLP ~~(``base model'') & 0.6359 \\ 
% \rule{0pt}{3ex}%
\hline
\textit{base model + label post-processing} & \textbf{0.6640} \\
\bottomrule
\end{tabular}
    \caption{Different embeddings, feature combinations and models for the development set of the TC task.
    % `Pre-softmax' refers to the pre-softmax layer of the linear classifier.
    Where the tokens for the BERT embeddings are not specified, all token embeddings are used as input to the linear classifier, whose pre-softmax layer of the linear classifier is referred to as `pre-softmax.'
    The results are mean values across five runs.
    The configuration for our final model is in italics.
    The full version of this table can be found in the appendix.
    }
    \label{tab:task2-ablation-abridged}
\end{table}
% \begin{table}[t]
%     \centering
%     \input{tables/task2-ablation}
%     \caption{Different embeddings, feature combinations and models for the development set of task 2.
%     Where the tokens for the BERT embeddings are not specified, all token embeddings are used as input for the linear classifier, whose pre-softmax layer of the linear classifier is referred to as
%     `pre-softmax.'
%     The results are mean values across five runs.
%     The configuration for our final model is in italics.}
%     \label{tab:task2-ablation}
% \end{table}

\subsection{Technique classification}

Our final model for the TC task achieves a 66.42\% F1-score on the development set and 57.37\% on the
test set, placing us eighth out of 31 teams.
Table~\ref{tab:task2-ablation-abridged} presents part of our feature ablation study; the full table can be found in the appendix.

We notice that our model design choices influence the model's performance at all stages of our system architecture.
Firstly, when choosing embeddings to represent the input fragments, we observe that, like in the SI task, the uncased BERT embeddings yield better results than the cased versions.
We also note that feeding the output of \texttt{BertForSequenceClassification}'s pre-softmax layer into another machine learning model leads to a higher F1-score than feeding in BERT embeddings (of \texttt{[CLS]} and the first ten tokens) directly: ca. 63\% versus 31.89-58.7\%.
The latter range is so large because the model choice for that set-up matters: the KimCNN yields significantly better results than a standard CNN, but it is still outperformed by the multilayer perceptron. 

Furthermore, we observe that the repetition pre-processing step markedly improves final results.
In the base model (sans additional features), adding this step boosts the overall F1-score by 7 percentage points to 63.41\%, and the \textsc{repetition} F1-score alone from 12.36 to 65.5\%.
We see that the choice of machine learning model has a modest effect for this architecture:
Using an SVM or a multilayer perceptron yields marginally better results than using the linear classifier whose layer is fed into the other models, which in turn outperforms the decision trees with extreme gradient boosting and the single-layer perceptron slightly.

Adding additional features also has a slight effect on the outcome.
% , either marginally decreasing or improving the overall F1-score.
Based on mean values across five model initializations, 
the bag-of-words features (\textit{America}, \textit{reductio}), the emotion feature and NE-2 slightly decrease the overall F1-score,
the sequence length feature and NE-6 do not change the result,
and the repetition count, the rhetorical phrase lexicon and the question mark feature as well as a combination of NE-2/6, the rhetorical lexicon and the question mark feature improve the score.
Interestingly, combining the features does not have an additive effect; some features work better in combination with others than on their own and vice versa.
Both the question mark feature on its own as well as together with NE-2 and the rhetorical lexicon score highest (63.59\%), but the scores of individual runs with the multi-feature model are more stable.
Further combinations of features are omitted from the table, but lead to results in the same range as the presented features and score lower than the base model set-up.

% Lastly, repetition post-processing 
% helps to boost model performance by almost 3 F1 percentage points beyond the base model, allowing our system to reach an F1 score of 66.40\%.

The propaganda technique we paid the most attention to while building the system, both during pre- and post-processing, is \textsc{repetition}.
This allows our team to achieve the best development phase result among all teams for \textsc{repetition} predictions (73.3\%).
The repetition post-processing steps
helps to boost model performance by almost 3 F1 percentage points beyond the base model, allowing our system to reach an overall F1 score of 66.40\%.

\subsection{Model performance by propaganda technique}

\begin{table}[t]
\centering
\renewcommand\arraystretch{1.1}
\begin{tabular}{@{}l @{\hskip -4mm } S[table-format=3.1] @{\hskip 1cm }  S[table-format=2.1] @{\hskip 1cm } S[table-format=2.1] S[table-format=2.1] S[table-format=2.1,table-space-text-pre={\textminus}~]@{}}
\toprule
&  & \multicolumn{1}{l}{~SI} & \multicolumn{3}{c}{TC} \\ 
\cmidrule(r{1cm}){3-3}\cmidrule{4-6}
\textbf{Technique}
& \textbf{\begin{tabular}[c]{@{}c@{}}Proportion\\(dev)\end{tabular}}
& \textbf{\begin{tabular}[c]{@{}c@{}}Recall\\(dev)\end{tabular}}
& \textbf{\begin{tabular}[c]{@{}c@{}}F1-score\\(dev)\end{tabular}}
& \textbf{\begin{tabular}[c]{@{}c@{}}F1-score\\(test)\end{tabular}}
& \textbf{Change} \\
\midrule
Loaded language & 30.6  & 70.6    & 76.6     & 74.7     & {\textminus}~1.9  \\
Name calling, labeling & 17.2 & 63.0     & 81.0   & 70.9  & {\textminus}~10.1\\
Repetition & 13.6  & 63.8     & 73.3    & 47.7  & {\textminus}~25.6 \\
Flag-waving & 8.2   & 74.4      & 73.7   & 54.4  & {\textminus}~19.3 \\
Exaggeration, minimisation & 6.4 & 57.6  & 52.7     & 28.3  & {\textminus}~24.4 \\
Doubt & 6.2 & 46.9  & 53.8    & 58.7     & {\MVPlus}~4.9   \\
Appeal to fear/prejudice & 4.4 & 62.9  & 30.6    & 39.9     & {\MVPlus}~9.3 \\
Slogans & 3.7 & 74.6 & 51.4     & 39.4  & {\textminus}~12.0   \\
Whataboutism, straw men, red herring & 2.7 & 36.8 & 0.0    & 0.0     & 0.0\\
Black-and-white fallacy & 2.1  & 46.9 & 21.4     & 23.7  & {\MVPlus}~2.3 \\
Causal oversimplification & 1.7 & 50.7 & 21.1     & 15.4  & {\textminus}~5.7  \\
Thought-terminating clich\'es & 1.6  & 51.4 & 17.4    & 23.8     & {\MVPlus}~6.4   \\
Appeal to authority & 1.3 & 49.9 & 18.2     & 14.7  & {\textminus}~3.5  \\
Bandwagon, reductio ad hitlerum& 0.5 & 8.4 & 22.2     & 12.2  & {\textminus}~10.0   \\
\midrule
All classes    & 100.0 & 63.8 & 66.4     & 57.4  & {\textminus}~9.0 \\
\bottomrule
\end{tabular}

\caption{
Technique-level breakdown of model performances for both subtasks.
The proportions, recall values and F1-scores are percentages.
The change of the F1-score
%from the development to the test phase 
is given in percentage points.
}
\label{tab:class-breakdown}
\end{table}

Our models do not perform equally well for each propaganda technique.
The scoreboard for the TC subtask includes F1-scores for each technique, and
with the help of the gold-standard labels for the development data for the TC subtask, we can calculate the recall score for each propaganda technique.
Table~\ref{tab:class-breakdown} shows this technique-level breakdown for both subtasks.

In the SI subtask,
the largest classes all have high recall scores (at least 63\%), likely due to their being well-represented in the data and because they tend to be very short text fragments (making it more probable to retrieve (nearly) complete spans).

The results for the smaller classes are mixed.
The recall score for \textsc{whataboutism, straw men, red herring} is relatively low,
and our technique classification system does not produce any predictions for the this class.
This might be due to the fact that such derailment techniques tend to be based on the discourse structure rather than more local syntactic or semantic patterns.% \cite{lewinski2013and}.

% CONFIRM?
% slogans and flag-waving: similar vocabulary -> boost

Our technique classification system also demonstrates a clear tendency towards better predictions for the classes representing large proportions of the training data,
which is expected because these classes are weighted more heavily in the micro-averaged F1 metric.
Each of three most frequent classes (\textsc{loaded language}, \textsc{name calling, labeling} and \textsc{repetition}) reaches an F1-score of above 70\% during the development phase (Table~\ref{tab:class-breakdown}). 

Most moderately frequent classes (\textsc{doubt}, \textsc{exaggeration, minimisation}, \textsc{flag-waving}, and \textsc{slogans}) score at least 50\% F1 in each category.
The only exception is for predictions on \textsc{appeal to fear/prejudice}.
% This class, however, does fare better in runs of our system on test data. %how much better?
Despite being a frequent class, predictions here top out at an F1 score of 30.6\% (on the development set).
As mentioned above, there are no predictions for \textsc{whataboutism, straw men, red herring}, but 
the five least frequent classes reach F1-scores of $20\pm3$\% during the development phase.

The confusion matrix of the development set predictions for the TC subtask can be found in the appendix (Figure~\ref{fig:task2-confusion-matrix}).
Quite often, the model misclassifies instances of other classes as \textsc{loaded language}, the most frequent class in the data.
For example, 36\% of \textsc{appeal to fear/prejudice} instances were classified as \textsc{loaded language}.
% Other interesting findings are the tendencies of the model to classify instances of "Black-and-white fallacy" as "Appeal to fear/prejudice", and "Thought-terminating cliches" as "Exaggeration, minimisation".
% NEED TO CONFIRM
% In the training data, many instances of "Flag-Waving" and "Slogans" share similar vocabulary, but this, interestingly, does not pose an issue for our system.

The model performs worse on test data, presumably due to having been overtuned on the development set or due to different label distributions in the development and test sets.
Only four classes manage to achieve better results in the test phase.
The most significant improvement is achieved in the \textsc{appeal to fear/prejudice} class, with an F1-score increase of more than 9 percentage points.
The most frequent class, \textsc{loaded language}, remains stable in test runs;
its F1-score decreases only by 1.9 percentage points.
The individual F1-score of 5 classes drop by more than 10 percentage points.
All those classes are quite frequent and represent almost half (49.1\%) of the instances of the development data.
As a result, the overall F1-score on test data drops by 9 percentage points.

The most significant decrease occurs in the \textsc{repetition} class with a drop of over 25 percentage points.
All teams demonstrate significantly worse test phase results for this technique compared to the development phase.
Only two teams out of 31 manage to score at least 50\% for this class, while only five more teams manage results above 30\%.

\section{Conclusion}

% A few summary sentences about your system, results, and ideas for future work.

While fair results on propaganda detection can be achieved with BERT embeddings alone, we further improve the performance on this task through the addition of linguistic features and pre- and post-processing techniques.
Our error analysis indicates that majority voting across several runs additionally increases the F1-score and stability of the model.

Future work can proceed in different directions.
In our experiments for the SI subtask, we applied majority voting across different runs of the same model,
but it could be beneficial 
to use majority voting
to combine predictions from the high-recall GloVe-based model with the high-precision BERT-based model.
Another potential source of improvement for either subtask is retraining the BERT model, similarly to \newcite{mapes-etal-2019-divisive} and \newcite{yoosuf2019fine}.
Finally, training a joint system for both subtasks, 
as \newcite{EMNLP19DaSanMartino} have done, 
may result in more accurate predictions.

\section*{Acknowledgments}
% Anyone you wish to thank who is not an author, which may include grants and anonymous reviewers.
% We would like to thank our professor  \c{C}a\u{g}r\i{} \c{C}{\"o}ltekin (Seminar f{\"u}r Sprachwissenschaft,
% Universit{\"a}t T{\"u}bingen) for his guidance and support throughout this project.

We thank Dr. \c{C}a\u{g}r\i{} \c{C}{\"o}ltekin for useful discussions and his guidance throughout this project.
%%% Looking at other papers, it seems quite common to skip the affiliation (but you include the academic title).

\bibliographystyle{coling}
\bibliography{semeval2020}

\begin{thebibliography}{}

\bibitem[\protect\citename{Al-Omari \bgroup et al.\egroup
  }2019]{al-omari-etal-2019-justdeep}
Hani Al-Omari, Malak Abdullah, Ola AlTiti, and Samira Shaikh.
\newblock 2019.
\newblock {JUSTD}eep at {NLP}4{IF} 2019 task 1: Propaganda detection using
  ensemble deep learning models.
\newblock In {\em Proceedings of the Second Workshop on Natural Language
  Processing for Internet Freedom: Censorship, Disinformation, and Propaganda},
  pages 113--118, Hong Kong, China, November. Association for Computational
  Linguistics.

\bibitem[\protect\citename{Alhindi \bgroup et al.\egroup
  }2019]{alhindi-etal-2019-fine}
Tariq Alhindi, Jonas Pfeiffer, and Smaranda Muresan.
\newblock 2019.
\newblock Fine-tuned neural models for propaganda detection at the sentence and
  fragment levels.
\newblock In {\em Proceedings of the Second Workshop on Natural Language
  Processing for Internet Freedom: Censorship, Disinformation, and Propaganda},
  pages 98--102, Hong Kong, China, November. Association for Computational
  Linguistics.

\bibitem[\protect\citename{Baccianella \bgroup et al.\egroup
  }2010]{baccianella2010sentiwordnet}
Stefano Baccianella, Andrea Esuli, and Fabrizio Sebastiani.
\newblock 2010.
\newblock Sentiwordnet 3.0: An enhanced lexical resource for sentiment analysis
  and opinion mining.
\newblock In {\em Lrec}, volume~10, pages 2200--2204.

\bibitem[\protect\citename{Barr{\'o}n-Cedeno \bgroup et al.\egroup
  }2019]{barron2019proppy}
Alberto Barr{\'o}n-Cedeno, Giovanni Da~San~Martino, Israa Jaradat, and Preslav
  Nakov.
\newblock 2019.
\newblock Proppy: A system to unmask propaganda in online news.
\newblock In {\em Proceedings of the AAAI Conference on Artificial
  Intelligence}, volume~33, pages 9847--9848.

\bibitem[\protect\citename{Bird \bgroup et al.\egroup }2009]{bird2009natural}
Steven Bird, Ewan Klein, and Edward Loper.
\newblock 2009.
\newblock {\em Natural language processing with {P}ython: Analyzing text with
  the {N}atural {L}anguage {T}oolkit}.
\newblock O'Reilly Media, Inc.

\bibitem[\protect\citename{Chen and Guestrin}2016]{chen2016xgboost}
Tianqi Chen and Carlos Guestrin.
\newblock 2016.
\newblock {XGB}oost: A scalable tree boosting system.
\newblock In {\em Proceedings of the 22nd acm sigkdd international conference
  on knowledge discovery and data mining}, pages 785--794.

\bibitem[\protect\citename{Da~San~Martino \bgroup et al.\egroup
  }2019a]{da2019findings}
Giovanni Da~San~Martino, Alberto Barron-Cedeno, and Preslav Nakov.
\newblock 2019a.
\newblock Findings of the {NLP4IF}-2019 shared task on fine-grained propaganda
  detection.
\newblock In {\em Proceedings of the Second Workshop on Natural Language
  Processing for Internet Freedom: Censorship, Disinformation, and Propaganda},
  pages 162--170.

\bibitem[\protect\citename{Da~San~Martino \bgroup et al.\egroup
  }2019b]{EMNLP19DaSanMartino}
Giovanni Da~San~Martino, Seunghak Yu, Alberto Barr\'{o}n-Cede\~no, Rostislav
  Petrov, and Preslav Nakov.
\newblock 2019b.
\newblock Fine-grained analysis of propaganda in news articles.
\newblock In {\em Proceedings of the 2019 Conference on Empirical Methods in
  Natural Language Processing and the 9th International Joint Conference on
  Natural Language Processing, EMNLP-IJCNLP 2019}, EMNLP-IJCNLP 2019, Hong
  Kong, China, November.

\bibitem[\protect\citename{Da~San~Martino \bgroup et al.\egroup
  }2020]{DaSanMartinoSemeval20task11}
Giovanni Da~San~Martino, Alberto Barr\'{o}n-Cede\~no, Henning Wachsmuth,
  Rostislav Petrov, and Preslav Nakov.
\newblock 2020.
\newblock {SemEval}-2020 task 11: Detection of propaganda techniques in news
  articles.
\newblock In {\em Proceedings of the 14th International Workshop on Semantic
  Evaluation}, SemEval 2020, Barcelona, Spain, December.

\bibitem[\protect\citename{Devlin \bgroup et al.\egroup }2019]{devlin2019bert}
Jacob Devlin, Ming-Wei Chang, Kenton Lee, and Kristina Toutanova.
\newblock 2019.
\newblock {BERT}: Pre-training of deep bidirectional transformers for language
  understanding.
\newblock In {\em Proceedings of the 2019 Conference of the North {A}merican
  Chapter of the Association for Computational Linguistics: Human Language
  Technologies, Volume 1 (Long and Short Papers)}, pages 4171--4186,
  Minneapolis, Minnesota, June. Association for Computational Linguistics.

\bibitem[\protect\citename{Ferreira~Cruz \bgroup et al.\egroup
  }2019]{ferreira-cruz-etal-2019-sentence}
Andr{\'e} Ferreira~Cruz, Gil Rocha, and Henrique Lopes~Cardoso.
\newblock 2019.
\newblock On sentence representations for propaganda detection: From
  handcrafted features to word embeddings.
\newblock In {\em Proceedings of the Second Workshop on Natural Language
  Processing for Internet Freedom: Censorship, Disinformation, and Propaganda},
  pages 107--112, Hong Kong, China, November. Association for Computational
  Linguistics.

\bibitem[\protect\citename{Graves and Schmidhuber}2005]{graves2005framewise}
Alex Graves and J{\"u}rgen Schmidhuber.
\newblock 2005.
\newblock Framewise phoneme classification with bidirectional {LSTM} and other
  neural network architectures.
\newblock {\em Neural networks}, 18(5-6):602--610.

\bibitem[\protect\citename{Gupta \bgroup et al.\egroup
  }2019]{gupta-etal-2019-neural}
Pankaj Gupta, Khushbu Saxena, Usama Yaseen, Thomas Runkler, and Hinrich
  Sch{\"u}tze.
\newblock 2019.
\newblock Neural architectures for fine-grained propaganda detection in news.
\newblock In {\em Proceedings of the Second Workshop on Natural Language
  Processing for Internet Freedom: Censorship, Disinformation, and Propaganda},
  pages 92--97, Hong Kong, China, November. Association for Computational
  Linguistics.

\bibitem[\protect\citename{Jowett and O'Donnell}2019]{jowett2019propaganda}
G.~Jowett and V.~O'Donnell.
\newblock 2019.
\newblock {\em Propaganda and Persuasion}.
\newblock Sage, 7th edition.

\bibitem[\protect\citename{Kim}2014]{kim-2014-convolutional}
Yoon Kim.
\newblock 2014.
\newblock Convolutional neural networks for sentence classification.
\newblock In {\em Proceedings of the 2014 Conference on Empirical Methods in
  Natural Language Processing ({EMNLP})}, pages 1746--1751, Doha, Qatar,
  October. Association for Computational Linguistics.

\bibitem[\protect\citename{Li \bgroup et al.\egroup
  }2019]{li-etal-2019-detection}
Jinfen Li, Zhihao Ye, and Lu~Xiao.
\newblock 2019.
\newblock Detection of propaganda using logistic regression.
\newblock In {\em Proceedings of the Second Workshop on Natural Language
  Processing for Internet Freedom: Censorship, Disinformation, and Propaganda},
  pages 119--124, Hong Kong, China, November. Association for Computational
  Linguistics.

\bibitem[\protect\citename{Mapes \bgroup et al.\egroup
  }2019]{mapes-etal-2019-divisive}
Norman Mapes, Anna White, Radhika Medury, and Sumeet Dua.
\newblock 2019.
\newblock Divisive language and propaganda detection using multi-head attention
  transformers with deep learning {BERT}-based language models for binary
  classification.
\newblock In {\em Proceedings of the Second Workshop on Natural Language
  Processing for Internet Freedom: Censorship, Disinformation, and Propaganda},
  pages 103--106, Hong Kong, China, November. Association for Computational
  Linguistics.

\bibitem[\protect\citename{Pedregosa \bgroup et al.\egroup }2011]{scikit-learn}
F.~Pedregosa, G.~Varoquaux, A.~Gramfort, V.~Michel, B.~Thirion, O.~Grisel,
  M.~Blondel, P.~Prettenhofer, R.~Weiss, V.~Dubourg, J.~Vanderplas, A.~Passos,
  D.~Cournapeau, M.~Brucher, M.~Perrot, and E.~Duchesnay.
\newblock 2011.
\newblock Scikit-learn: Machine learning in {P}ython.
\newblock {\em Journal of Machine Learning Research}, 12:2825--2830.

\bibitem[\protect\citename{Pennebaker \bgroup et al.\egroup
  }2001]{pennebaker2001linguistic}
James~W Pennebaker, Martha~E Francis, and Roger~J Booth.
\newblock 2001.
\newblock Linguistic inquiry and word count: {LIWC} 2001.
\newblock {\em Mahway: Lawrence Erlbaum Associates}, 71(2001):2001.

\bibitem[\protect\citename{Pennington \bgroup et al.\egroup
  }2014]{pennington2014glove}
Jeffrey Pennington, Richard Socher, and Christopher~D. Manning.
\newblock 2014.
\newblock Glo{V}e: Global vectors for word representation.
\newblock In {\em Empirical Methods in Natural Language Processing (EMNLP)},
  pages 1532--1543.

\bibitem[\protect\citename{Rashkin \bgroup et al.\egroup
  }2017]{rashkin2017truth}
Hannah Rashkin, Eunsol Choi, Jin~Yea Jang, Svitlana Volkova, and Yejin Choi.
\newblock 2017.
\newblock Truth of varying shades: Analyzing language in fake news and
  political fact-checking.
\newblock In {\em Proceedings of the 2017 Conference on Empirical Methods in
  Natural Language Processing}, pages 2931--2937.

\bibitem[\protect\citename{Somasundaran \bgroup et al.\egroup
  }2007]{somasundaran2007detecting}
Swapna Somasundaran, Josef Ruppenhofer, and Janyce Wiebe.
\newblock 2007.
\newblock Detecting arguing and sentiment in meetings.
\newblock In {\em Proceedings of the SIGdial Workshop on Discourse and
  Dialogue}, volume~6.

\bibitem[\protect\citename{Wolf \bgroup et al.\egroup
  }2019]{Wolf2019HuggingFacesTS}
Thomas Wolf, Lysandre Debut, Victor Sanh, Julien Chaumond, Clement Delangue,
  Anthony Moi, Pierric Cistac, Tim Rault, R'emi Louf, Morgan Funtowicz, and
  Jamie Brew.
\newblock 2019.
\newblock Huggingface's transformers: State-of-the-art natural language
  processing.
\newblock {\em ArXiv}, abs/1910.03771.

\bibitem[\protect\citename{Yoosuf and Yang}2019]{yoosuf2019fine}
Shehel Yoosuf and Yin Yang.
\newblock 2019.
\newblock Fine-grained propaganda detection with fine-tuned {BERT}.
\newblock In {\em Proceedings of the Second Workshop on Natural Language
  Processing for Internet Freedom: Censorship, Disinformation, and Propaganda},
  pages 87--91.

\end{thebibliography}

\newpage

% \appendix
\begin{appendices}
% \makeatletter
% \def\@seccntformat{\appendixname\ %
% \csname thesection\endcsname. \quad}
% \makeatother
\section{Supplemental Material}
\label{sec:supplemental}

% Any low-level implementation details—rules and pre-/post-processing steps, features, hyperparameters, etc.—that would help the reader to replicate your system and experiments, but are not necessary to understand major design points of the system and experiments.

% Any figures or results that aren’t crucial to the main points in your paper but might help an interested reader delve deeper.

\begin{table}[h]
    \centering
    \renewcommand\arraystretch{1.1}
\newcolumntype{R}{>{\raggedleft\arraybackslash}p{1.5cm}}
\begin{tabular}{@{} l R @{}}
\toprule
\textbf{Configuration}     & \textbf{F1-score}     \\
\midrule
\textsc{bert}-base-cased + pre-softmax + multilayer perceptron (MLP) & 0.5485\\
\textsc{bert}-base-uncased + pre-softmax + MLP & 0.5635
\\ % NO REPETITION PREPROCESSING
% \rule{0pt}{3ex}%
\hline
\textsc{bert}-base-uncased embeddings of \texttt{[CLS]} \& the first 10 tokens + CNN & 0.3189\\
\textsc{bert}-base-uncased embeddings of \texttt{[CLS]} \& the first 10 tokens + KimCNN & 0.4835\\
\textsc{bert}-base-uncased embeddings of \texttt{[CLS]} \& the first 10 tokens + MLP & 0.5870\\
% \rule{0pt}{3ex}%
% \cmidrule[0.02em]{1-2}
\hline
repetition pre-processing (rep) + \textsc{bert}-base-uncased + linear classifier & 0.6322 \\ 
rep + \textsc{bert}-base-uncased + pre-softmax + XGBoost  & 0.6219 \\
rep + \textsc{bert}-base-uncased + pre-softmax + single-layer perceptron & 0.6312  \\
rep + \textsc{bert}-base-uncased + pre-softmax + SVM & 0.6341 \\
rep + \textsc{bert}-base-uncased + pre-softmax + MLP & 0.6341 \\
% \rule{0pt}{3ex}%
\hline
rep + \textsc{bert}-base-uncased + pre-softmax + America + MLP & 0.6312 \\
rep + \textsc{bert}-base-uncased + pre-softmax + reductio + MLP & 0.6322 \\
rep + \textsc{bert}-base-uncased + pre-softmax + emotion + MLP & 0.6331 \\
rep + \textsc{bert}-base-uncased + pre-softmax + sequence length + MLP & 0.6341 \\
rep + \textsc{bert}-base-uncased + pre-softmax + repetition count + MLP & 0.6350 \\
% \rule{0pt}{3ex}%
\hline
rep + \textsc{bert}-base-uncased + pre-softmax + two named entity classes (\textsc{ne-2}) + MLP  & 0.6322\\ 
rep + \textsc{bert}-base-uncased + pre-softmax + six named entity classes (\textsc{ne-6}) + MLP  & 0.6341\\
rep + \textsc{bert}-base-uncased + pre-softmax + Arguing Lexicon (\textsc{al}) + MLP & 0.6350 \\
rep + \textsc{bert}-base-uncased + pre-softmax + question mark feature (\textsc{q}) + MLP  & 0.6359\\
rep + \textsc{bert}-base-uncased + pre-softmax + \textsc{ne-6} + \textsc{al} + \textsc{q} + MLP & 0.6350 \\
rep + \textsc{bert}-base-uncased + pre-softmax + \textsc{ne-2} + \textsc{al} + \textsc{q} + MLP ~~(``base model'') & 0.6359 \\ 
% \rule{0pt}{3ex}%
\hline
\textit{base model + label post-processing} & \textbf{0.6640} \\
\bottomrule
\end{tabular}
    \caption{Different embeddings, feature combinations and models for the development set of the TC task.
    Where the tokens for the BERT embeddings are not specified, all token embeddings are used as input to the linear classifier, whose pre-softmax layer of the linear classifier is referred to as
    `pre-softmax.'
    The results are mean values across five runs.
    The configuration for our final model is in italics.}
    \label{tab:task2-ablation}
\end{table}

\begin{figure}[h]
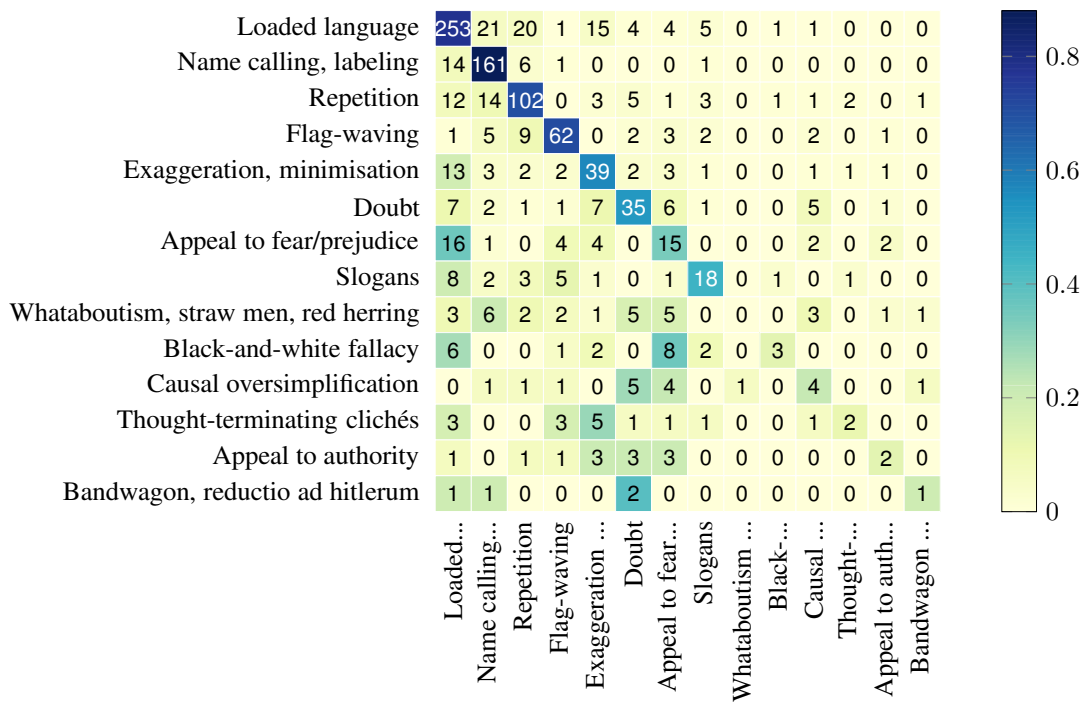

  \centering
    \includestandalone[width=0.9\textwidth]{figures/confusion-matrix}
  \caption{
  Confusion matrix for technique classification predictions on the development set.
  Rows represent true labels and columns predicted labels.
  The numbers in the matrix are absolute instance counts.
  The colour scale indicates each classification's frequency relative to its row (true label).
  }
   \label{fig:task2-confusion-matrix}
\end{figure}
\end{appendices}

\end{document}